\documentclass[runningheads]{llncs}
\usepackage[T1]{fontenc}
\usepackage{graphicx}
\usepackage[sortcites]{biblatex}
\usepackage{amsmath}
\usepackage{multirow}
\usepackage{fixltx2e}

\usepackage{url}
\usepackage{hyperref}
\usepackage[hyphenbreaks]{breakurl}

\newcommand{\beginsupplement}{%
        \setcounter{table}{0}
        \renewcommand{\thetable}{S\arabic{table}}%
        \setcounter{figure}{0}
        \renewcommand{\thefigure}{S\arabic{figure}}%
     }

\begin{document}

\title{Panoptic Segmentation of Mammograms with Text-To-Image Diffusion Model}

\newcommand{\repeatthanks}{\textsuperscript{\thefootnote}}

\author{Kun Zhao\thanks{These authors contributed equally to this work.}\inst{1} \and
Jakub Prokop\repeatthanks\inst{1} \and
Javier Montalt Tordera\inst{1} \and Sadegh Mohammadi\inst{1}}

\authorrunning{K. Zhao et al.}
\titlerunning{Panoptic Segmentation of Mammograms}

\institute{Bayer AG, Germany\\
\email{jakub.prokop@bayer.com}
}

\maketitle

\begin{abstract}
Mammography is crucial for breast cancer surveillance and early diagnosis. However, analyzing mammography images is a demanding task for radiologists, who often review hundreds of mammograms daily, leading to overdiagnosis and overtreatment. Computer-Aided Diagnosis (CAD) systems have been developed to assist in this process, but their capabilities, particularly in lesion segmentation, remained limited. With the contemporary advances in deep learning their performance may be improved. Recently, vision-language diffusion models emerged, demonstrating outstanding performance in image generation and transferability to various downstream tasks. We aim to harness their capabilities for breast lesion segmentation in a panoptic setting, which encompasses both semantic and instance-level predictions. Specifically, we propose leveraging pretrained features from a Stable Diffusion model as inputs to a state-of-the-art panoptic segmentation architecture, resulting in accurate delineation of individual breast lesions. To bridge the gap between natural and medical imaging domains, we incorporated a mammography-specific MAM-E diffusion model and BiomedCLIP image and text encoders into this framework. We evaluated our approach on two recently published mammography datasets, CDD-CESM and VinDr-Mammo. For the instance segmentation task, we noted 40.25 AP\textsubscript{0.1} and 46.82 AP\textsubscript{0.05}, as well as 25.44 PQ\textsubscript{0.1} and 26.92 PQ\textsubscript{0.05}. For the semantic segmentation task, we achieved Dice scores of 38.86 and 40.92, respectively.

\end{abstract}

\tolerance=1
\emergencystretch=\maxdimen
\hyphenpenalty=10000
\hbadness=10000

\section{Introduction}
\label{cha 1: Introduction}

Breast cancer is the most common cancer among women worldwide, with an estimated 2.3 million new cases diagnosed in 2020, accounting for 11.7\% of all cancer cases globally~\cite{sung2021global}. Early detection and accurate diagnosis, followed by timely treatment, are crucial for improving survival rates and patient outcomes. Mammography screening plays a pivotal role in early detection, reducing breast-cancer-related deaths by 15\% to 25\%~\cite{loberg2015benefits}. However, interpreting mammography images is difficult and tedious, prone to human error~\cite{guo2022review}, often resulting in false positive diagnoses and unnecessary biopsies~\cite{sung2021global, loberg2015benefits}. 

Computer-Aided Diagnosis (CAD) systems have been developed to enhance the diagnostic process~\cite{liew2021review}. These systems integrate detection and analysis capabilities, enabling them to identify and delineate abnormal tissue regions. Additionally, they assist radiologists by classifying suspicious lesions and offering valuable ``second opinions'', thus supporting more informed clinical decision-making~\cite{liew2021review, hassan2022mammogram, guo2022review}.

However, their performance still lags behind human expertise~\cite{loizidou2023computer}. Mammography image segmentation is particularly challenging due to the complexity and variability of breast tissue patterns, especially in dense breasts, which makes it difficult to distinguish between normal and abnormal areas~\cite{hassan2022mammogram, loizidou2023computer, abo2024advances}. Diagnosing foci and non-mass-like enhancing lesions is particularly challenging due to their small size and unclear boundaries~\cite{guo2022review}. Current research primarily focuses on analyzing masses and micro-calcifications~\cite{loizidou2023computer}, often overlooking other abnormalities such as architectural distortions, non-mass enhancements, and asymmetries. This challenge is further compounded by low contrast levels and high noise, which can obscure lesions~\cite{abo2024advances}.

To address these issues, mammogram segmentation research has been exploring two main directions: providing global information about malignant tissues through \textit{semantic segmentation}~\cite{yan2021two, hossain2022microc} and offering a more fine-grained diagnosis of individual lesions through \textit{instance segmentation}~\cite{bhatti2020multi, soltani2021breast, ahmed2023images}. A significant shortcoming of previous work is the consideration of semantic and instance segmentation as separate tasks. This underscores the importance of developing a unified framework that addresses both tasks simultaneously, enabling the integration of global information and detailed lesion-specific insights, ultimately enhancing diagnostic accuracy and effectiveness.

\textit{Panoptic segmentation} is an emerging image analysis method that unifies semantic and instance segmentation to provide a comprehensive understanding of visual scenes~\cite{kirillov2019panoptic}. This integration allows for more detailed and accurate image analysis, making it particularly useful for complex scenes where distinguishing between overlapping objects is crucial. By combining the strengths of both segmentation methods, panoptic segmentation offers a holistic view that significantly enhances the accuracy and reliability of image interpretation in various applications, including medical imaging ~\cite{chuang2023deep}. 

The rise of Diffusion Models~\cite{dhariwal2021diffusion, rombach2022high} and their integration into visual-language models promise further advancements. These architectures enhance robust feature extraction and rich semantic understanding, improving the ability to distinguish complex objects and patterns~\cite{kim2024eclipse}. A relevant example of this approach is ODISE (Open-Vocabulary Panoptic Segmentation with Text-to-Image Diffusion Models)~\cite{xu2023open}, which utilizes the Stable Diffusion model to integrate visual and textual information, achieving comprehensive and precise segmentation based on textual descriptions.

Inspired by ODISE~\cite{xu2023open}, we propose M-ODISE (Mammography Open-Vocabulary Panoptic Segmentation with Text-to-Image Diffusion Models), a novel panoptic segmentation approach designed to delineate a wide variety of breast lesions, ultimately aimed at improving cancer diagnosis and characterization.

Our contributions are as follows: \textit{i)}~To the best of our knowledge, M-ODISE is the first work to apply panoptic segmentation to mammography images.
\textit{ii)}~We evaluate the M-ODISE framework across several component choices, varying the image encoder, the text encoder, or both simultaneously.
\textit{iii)}~We evaluate the proposed method using two recently released mammography datasets: CDD-CESM~\cite{khaled2022categorized} and VinDr-Mammo~\cite{nguyen2023vindr}.

\nopagebreak
\section{Materials and methods}
\label{cha 3: Methods}

This section outlines our segmentation framework and evaluation procedure.

\subsection{Segmentation framework}

\begin{figure}[h]
\includegraphics[width=\columnwidth]{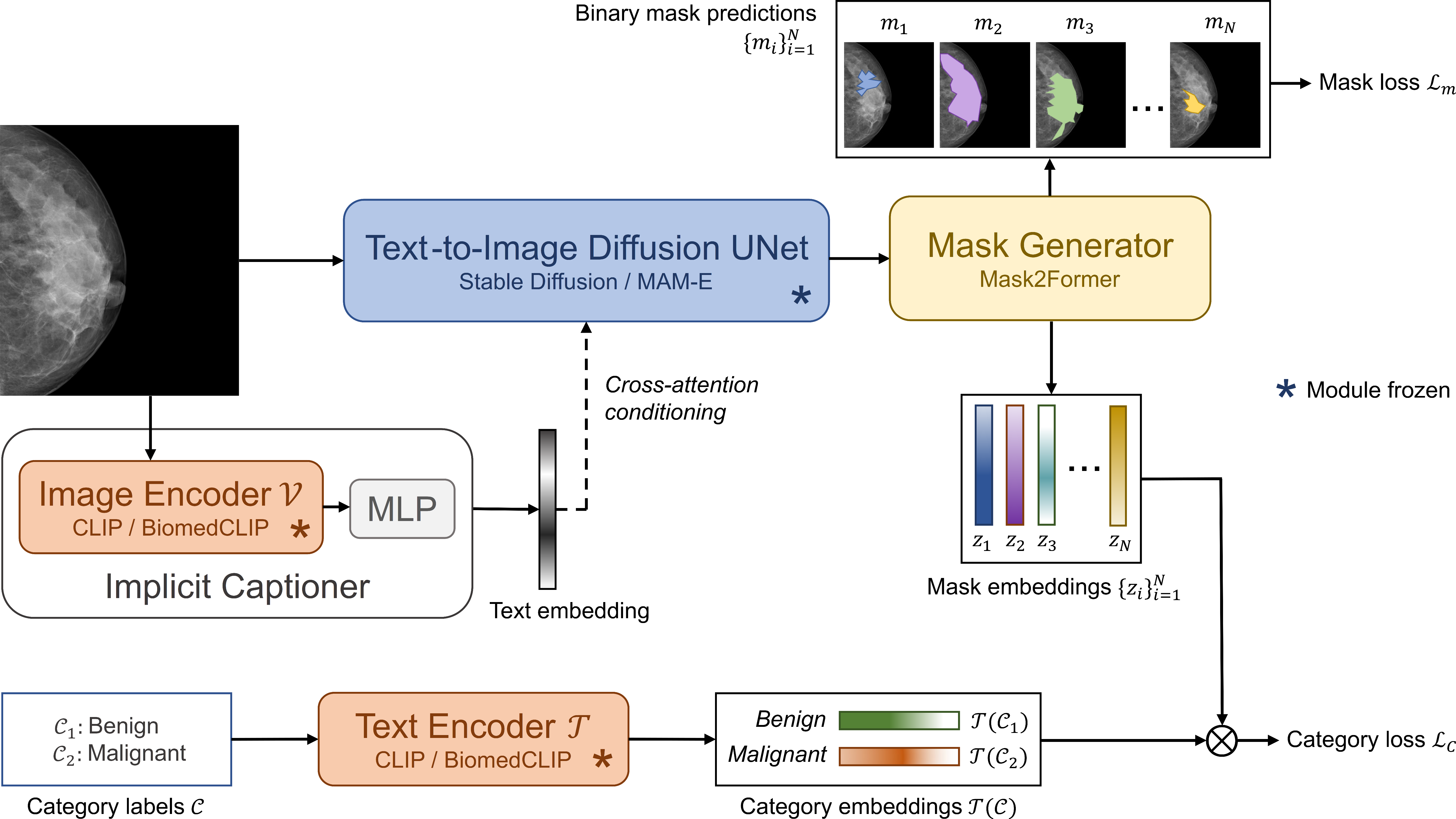}
\caption{
The overview of our framework, adapted from ODISE. 
Features extracted by the text-to-image diffusion model are passed to a mask generator, which outputs binary mask predictions and mask embeddings for individual objects detected in the image. These mask embeddings are then combined with category embeddings from the text encoder via a dot product to supervise the classification task. Additionally, an implicit captioner encodes the image to provide a conditioning signal for the diffusion process.}
\label{fig: fig1}
\end{figure}

In our approach, we directly adopted the ODISE framework, which consists of several modules, as illustrated in Figure~\ref{fig: fig1}. First, the \textit{implicit captioner} encodes an image into a vector of embeddings. These embeddings are then passed as a guidance signal to \textit{the text-to-image diffusion model}, which extracts a stack of feature maps from the image. The features are then fed into the \textit{mask generator} module, which detects objects and semantic areas in the image and produces segmentation maps. Finally, the \textit{classification head} generates the class prediction for each mask. Below, we provide a detailed overview of these modules. In our implementation, we used the publicly available code\footnote{https://github.com/NVlabs/ODISE} provided by the ODISE and Mask2Former authors, built on the detectron2 library~\cite{wu2019detectron2}.

\subsubsection{Text-to-image diffusion model.} 

To bridge the gap between natural and medical imaging domains, we replaced the Stable Diffusion (SD) feature extractor incorporated into ODISE \cite{xu2023open} with a mammography-specific MAM-E model~\cite{montoya2024mam}. MAM-E is based on SD architecture, trained on approximately 55,000 healthy mammography images from the OMI-H~\cite{halling2020optimam} and VinDr-Mammo~\cite{nguyen2023vindr} datasets, encompassing bilateral craniocaudal (CC) and mediolateral oblique (MLO) mammogram modalities.

During both training and inference we keep this model frozen. First, given an image $x$, we sample a noisy image $x_t$ at time step $t=0$ as:

\begin{equation} 
\label{eq_s}
x_t = \sqrt{\bar{\alpha_t}}x + \sqrt{1 - \bar{\alpha}_t }\epsilon, \quad \epsilon\sim \mathcal{N}(0, I),
\end{equation}

\noindent where $\alpha_0, ..., \alpha_{t}$ is a pre-defined noise schedule~\cite{ho2020denoising} and $\bar{\alpha}_{t} = \prod_{k=0}^{t}a_k$~\cite{xu2023open}. Then, we pass $x_t$ through the denoising UNet of MAM-E model and extract features from its intermediate layers, similarly to \cite{xu2023open}.

\subsubsection{Implicit captioner.} 

The implicit captioner is used to obtain the conditioning signal needed to guide the diffusion process through the cross-attention mechanism~\cite{xu2023open, rombach2022high}. The architecture of this module consists of two parts. First, a frozen image encoder $\mathcal{V}$ encodes the image into a vector of meaningful text embeddings. Then, a multilayer perceptron (MLP) layer is fine-tuned to project these embeddings into the space of the conditioning signal of the frozen diffusion model. The image encoder $\mathcal{V}$ is derived from BiomedCLIP~\cite{BiomedCLIP}, which is a CLIP architecture specially tailored for handling medical data. Specifically, it was trained using 15 million figure-caption pairs sourced from various biomedical research articles.

\subsubsection{Mask generator.} 

The features $f$ extracted by the diffusion model are subsequently fed into the mask generator~$\mathcal{M}_G$. Following \cite{xu2023open}, we implemented it using the Mask2Former architecture without its feature extractor backbone.

Mask2Former is a universal image segmentation framework, which, aside from the feature extractor, encompasses two modules. A pixel decoder upsamples the features into high-resolution pixel embeddings. A transformer decoder equipped with the masked-attention mechanism processes the learnable object queries representing different object instances~\cite{bhatti2020multi}. 

The output of this model consists of $N$ class-agnostic binary masks $\{m\}_{i=1}^{N}$ and $N$ mask embeddings $\{z_i\}_{i=1}^{N}$ that represent different objects detected in the images as follows:
\begin{equation}
      \{m\}_{i=1}^{N},\,\{z_i\}_{i=1}^{N} = \mathcal{M}_G(f, \, q),
\end{equation}
During training, the masks $\{m\}_{i=1}^{N}$ are compared against the ground-truth masks with the loss function $\mathcal{L}_m$ that combines binary cross-entropy loss and Dice loss~\cite{bhatti2020multi}.
Embeddings $\{z_i\}_{i=1}^N$ are further fed into the classification head to generate the class predictions. 

\subsubsection{Classification head.} 

In the classification head, the category labels $\mathcal{C}$ are encoded using the text encoder $\mathcal{T}$ to provide the categorical text embeddings $\mathcal{T}(\mathcal{C})$. The text embeddings and mask embeddings $\{z_i\}_{i=1}^{N}$ are then combined through a dot product and passed through a softmax function to produce class probabilities. The cross-entropy loss $\mathcal{L}_\mathcal{C}$ is subsequently computed to supervise the training process.

\subsection{Datasets}
We trained and evaluated the segmentation model using two mammography datasets. 

The CDD-CESM dataset~\cite{khaled2022categorized} consists of 2006 contrast-enhanced mammograms, categorized by a board of radiologists into 757 normal, 587 benign, and 662 malignant cases. Hand-drawn masks were provided for all abnormal findings. In our experiments, we processed only images with masks, excluding the normal cases.

VinDr-Mammo~\cite{nguyen2023vindr} is a large public full-field digital mammography dataset containing 10,000 pairs of MLO and CC images. Since benign findings are not annotated, we defined the object categories based on the available BI-RADS (Breast Imaging Reporting and Data System)~\cite{acr} score. 

VinDr-Mammo annotations encompass cases with scores ranging from 3 (probably benign) to 5 (probably malignant). Specifically, there are 465 BI-RADS~3, 381 BI-RADS~4, and 113 BI-RADS~5 cases with bounding boxes available. To adapt this dataset for panoptic segmentation, we generated panoptic masks by using an adaptive Otsu thresholding algorithm on the Gaussian blurred bounding box regions, with $\sigma = 7$. Subsequently, within each bounding box area, identified areas of interest were enclosed in one segment through the concave hull algorithm.

\subsection{Evaluation metrics}

We evaluated the model in both instance and semantic segmentation.
For semantic segmentation, the model's effectiveness was measured using the Dice coefficient, averaged across classes. 

In instance segmentation, performance is typically reported as the mean average precision (AP) with an intersection over union (IoU) threshold of 0.5~\cite{wu2019detectron2}. However, breast lesion masks vary significantly in size and shape, and radiologists often disagree on the precise ground-truth assignment~\cite{zbinden2023stochastic}. Given this, we believe the conventional IoU threshold of 0.5 might be too strict and investigate various threshold values to balance the number of detected lesions against segmentation accuracy.

To achieve this, we use metrics traditionally employed in panoptic segmentation, namely \textit{recognition quality} (RQ) and \textit{segmentation quality} (SQ)~\cite{kirillov2019panoptic}. RQ is defined as the F1 score over the number of true positive (TP), false positive (FP) and false negative (FN) object detections. A high RQ indicates that most ground-truth objects were detected. SQ, on the other hand, measures the correctness of the predicted mask and is defined as the average IoU over true positive object predictions. Their product is referred to as \textit{panoptic quality} (PQ)~\cite{kirillov2019panoptic}.

To determine the optimal IoU threshold, we computed PQ for IoU thresholds ranging from 0.05 to 0.9 in steps of 0.05 and selected the threshold with the highest PQ value. We then report PQ, RQ, SQ, and AP for the selected threshold. To enhance result reliability, we used 5-fold cross-validation. For the ablation study on threshold values please refer to the supplementary material.

\section{Results}
\label{cha 4: Results}

We present the results of M-ODISE in semantic and instance segmentation, comparing it to three baselines: Mask2Former (M2F) with Swin-L~\cite{liu2021swin} and Stable Diffusion (SD)~\cite{rombach2022high} backbones, as well as the ODISE model described in~\cite{xu2023open}. Additionally, we explore the impact of BiomedCLIP and MAM-E on the final model performance, demonstrating how these modules affect ODISE. Table~\ref{tab: results cdd cesm} and Table~\ref{tab: results vindr mammo} display the evaluation results on the CDD-CESM and VinDr-Mammo datasets, respectively. 

\begin{figure}[h]
\centering
\includegraphics[width=\textwidth, keepaspectratio]{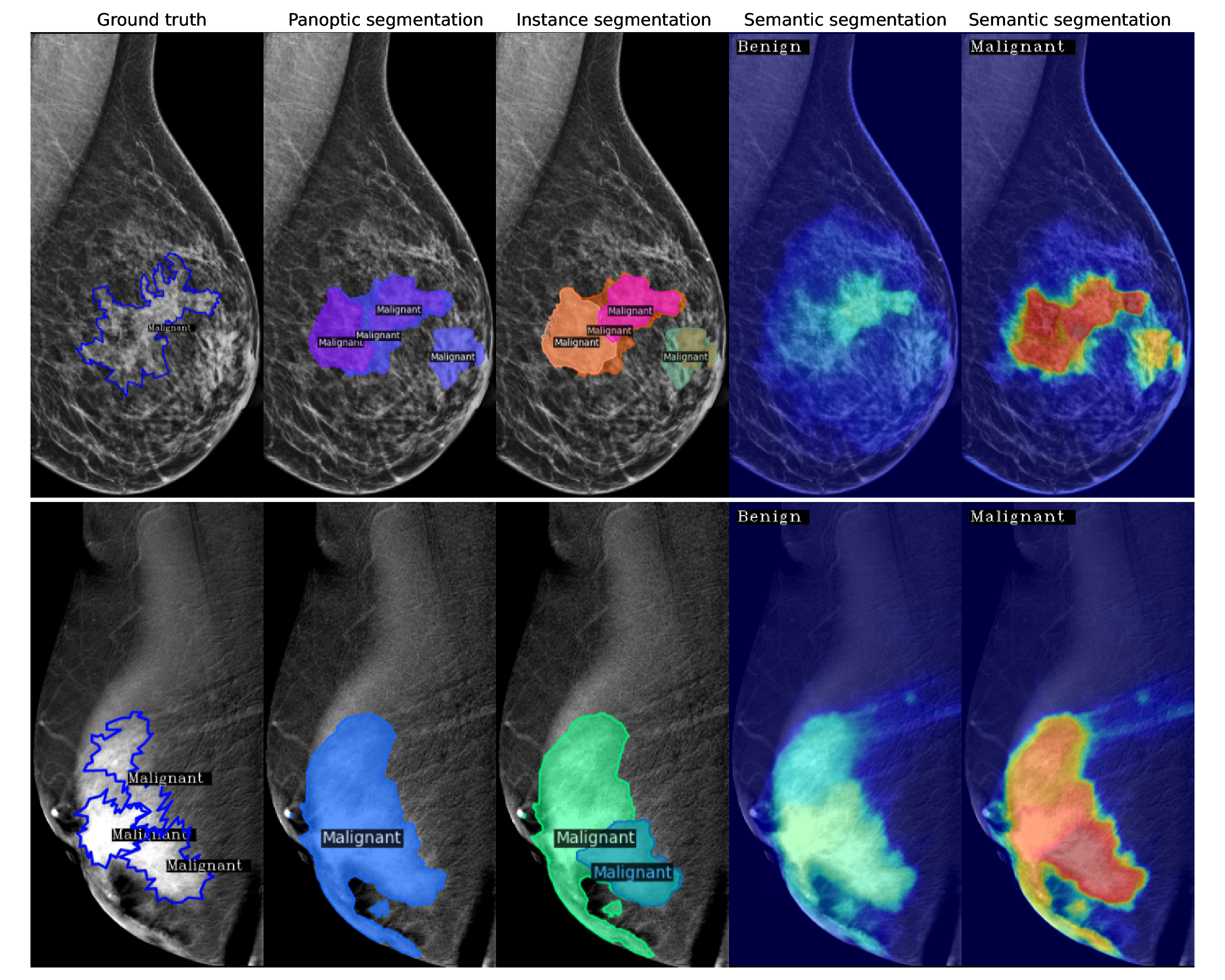}
\caption{
Qualitative visualization of M-ODISE predictions on the CDD-CESM dataset. For more visual examples please refer to the supplementary material. 
}
\label{fig: fig2}
\end{figure}

On the CDD-CESM dataset, M-ODISE marginally outperformed the baselines in RQ, PQ, and AP but lagged behind ODISE with the MAM-E model in SQ and behind both Mask2Former and ODISE with MAM-E in the Dice score. On the VinDr-Mammo dataset, M-ODISE did not surpass ODISE, ODISE with BiomedCLIP, or ODISE with MAM-E. The base ODISE achieved the best results in RQ, PQ, and Dice coefficient, ODISE with BiomedCLIP led in SQ, and ODISE with the MAM-E model had the highest AP value. On both datasets, Mask2Former with the SD backbone performed significantly worse than the other models across all metrics. 

Visual prediction examples are provided in Figure~\ref{fig: fig2}. M-ODISE successfully detected architectural distortion (top) and heterogeneously enhancing masses (bottom). In both cases, the model localized abnormal areas but struggled to differentiate between individual lesions. While its predictions included false positive lesions, the semantic heatmaps indicate that these false positives have lower probabilities compared to the true positives.

\begin{table}
\begin{center}
\scriptsize
\def\arraystretch{1.1}%
\setlength{\tabcolsep}{3pt}
\caption{Segmentation results on CDD-CESM dataset. The mean and standard deviation of each metric were gathered from 5-fold cross-validation. The highest mean values are marked in bold.}
\begin{tabular}{|l|c|c|c|c|c|c|c|c|}
\hline
Model & RQ\textsubscript{0.1} & SQ\textsubscript{0.1} & PQ\textsubscript{0.1} & AP\textsubscript{0.1} & Dice\\
\hline
M2F (Swin-L) & 45.29 \textpm 3.39 & 54.77 \textpm 3.16 & 24.65 \textpm 1.08 & 34.46 \textpm 0.68 & \textbf{40.11} \textpm 4.52 \\
M2F (SD) & 34.39 \textpm 4.28 & 51.04 \textpm 2.59 & 17.03 \textpm 2.77 & 29.02 \textpm 5.59 & 27.85 \textpm 6.35\\
ODISE & 43.70 \textpm  4.02 & 55.37 \textpm 1.88 & 24.09 \textpm 1.83 & 38.90 \textpm 3.29 & 36.92 \textpm 2.85\\
ODISE + BiomedCLIP & 44.06 \textpm 3.92 & 55.53 \textpm 1.00 & 24.33 \textpm 1.93 & 39.02 \textpm 3.74 & 37.50 \textpm 3.24\\
ODISE + MAM-E & 45.08 \textpm 3.36 & \textbf{55.75} \textpm 1.97 & 25.06 \textpm 1.75 & 39.75 \textpm 3.15 & 36.98 \textpm 2.88 \\
M-ODISE & \textbf{46.19} \textpm 3.39 & 55.26 \textpm 1.22 & \textbf{25.44} \textpm 1.87 & \textbf{40.25} \textpm 4.90 & 38.86 \textpm 2.77\\
\hline

\end{tabular}
\end{center}
\label{tab: results cdd cesm}
\end{table}

\begin{table}
\begin{center}
\scriptsize
\def\arraystretch{1.1}%
\setlength{\tabcolsep}{3pt}
\caption{Segmentation results on VinDr-Mammo dataset. Mean and standard deviation from 5-fold cross-validation are presented for each metric. The highest mean values are marked in bold.}
\begin{tabular}{|l|c|c|c|c|c|c|}
\hline

Model & RQ\textsubscript{0.05} & SQ\textsubscript{0.05} & PQ\textsubscript{0.05} & AP\textsubscript{0.05} & Dice\\
\hline
M2F (Swin-L) & 45.79 \textpm 4.49 & 49.35 \textpm 4.68 & 23.04 \textpm 2.36 & 44.77 \textpm 1.97 & 27.84 \textpm 1.41\\
M2F (SD) & 34.55 \textpm 3.36 & 52.42 \textpm 2.97 & 17.96 \textpm 2.56 & 39.11 \textpm 3.94 & 28.59 \textpm 7.49\\
ODISE & \textbf{47.66} \textpm 1.46 & 57.06 \textpm 1.98 & \textbf{27.40} \textpm 1.52 & 45.83 \textpm 2.25 & \textbf{41.67} \textpm 2.94\\
ODISE + BiomedCLIP & 46.84 \textpm 2.47 & \textbf{57.81} \textpm 2.07 & 27.22 \textpm 2.07 & 47.18 \textpm 2.93 & 41.49 \textpm 3.48\\
ODISE + MAM-E & 46.53 \textpm 2.56 & 56.61 \textpm 2.11 & 26.43 \textpm 1.83 & \textbf{47.76} \textpm 2.99 & 41.49  \textpm 5.73\\
M-ODISE & 46.26 \textpm 2.90 & 57.66 \textpm 1.27 & 26.92 \textpm 1.89 & 46.82 \textpm 2.97 & 40.92 \textpm 4.46\\
\hline
\end{tabular}
\end{center}
\label{tab: results vindr mammo}
\end{table}

\section{Discussion}
\label{cha 5: Discussion}

The results in Table~\ref{tab: results cdd cesm} and Table~\ref{tab: results vindr mammo} present ambiguous findings, showing that M-ODISE usually exceeds other ODISE variants in CDD-CESM segmentation by a small margin, but underperforms on VinDr-Mammo dataset. 
These observations suggest that the advantages of domain adaptations of foundational models for medical tasks are not definitive and may greatly depend on the specific model architecture, as evidenced in medical image classification studies~\cite{huix2024natural}.

In most scenarios, ODISE and M-ODISE outperform the base Mask2Former in instance segmentation in terms of AP, underscoring the potential of diffusion-based segmentation models. 
However, performance declines when the text encoder is omitted from the pipeline, as demonstrated by Mask2Former using the SD feature extractor. 
A plausible explanation for this phenomenon is proposed in \cite{xu2023open}, which notes that removing the supervision from label-text encoding may hinder the model's classification abilities.

In semantic segmentation, the advantage of diffusion-based segmentation models over the Mask2Former baseline is unclear, as their performance lags slightly behind in the case of CDD-CESM data. 
However, with the VinDr-Mammo images, the ODISE architectures prove to be significantly more effective, without compromising their instance segmentation capabilities. 

To our knowledge, we are the first to conduct joint semantic and instance segmentation on a variety of breast lesions.
Previous works by Bhatti et al.~\cite{bhatti2020multi} and Ahmed et al.~\cite{ahmed2023images} attempted similar instance segmentation tasks, reporting AP values of 0.84 and 0.75-0.80, respectively. 
Although their metric definitions differ from ours~\cite{wu2019detectron2}, it is also possible that the panoptic approach does not yet match the performance of specialized instance segmentation models in this domain. 
Similarly, the semantic segmentation capabilities of Mask2Former, ODISE, and M-ODISE lag behind those of architectures dedicated to specific semantic tasks, which often achieve over 0.90 Dice scores in narrowly defined lesion segmentation problems \cite{abo2024advances}. 
Considering this, we hypothesize that further technical advancements are needed to tailor current panoptic segmentation techniques to the specifics of mammogram analysis.

A limiting factor in our work, and breast cancer research in general, is the scarcity of mammography data~\cite{hassan2022mammogram}. 
Out of 20,000 images in VinDr-Mammo, the largest public mammography dataset, only 1,768 images have bounding box annotations, which are limited to malignant or suspicious areas, excluding benign tumors and other breast tissues~\cite{nguyen2023vindr}. 

Moreover, the available masks are not biopsy-confirmed, and the radiologist-subjective delineation of individual lesions poses a challenge for instance segmentation, as seen in Figure~\ref{fig: fig2}.
Finally, differences in annotation protocols hinder training on multiple data sources, negatively impacting the final model performance and applicability in diverse, real-world scenarios. 
The development of new, large-scale, public mammography datasets in the future could mitigate these limitations. 

In this study, we proposed M-ODISE, a panoptic segmentation framework for the delineation of breast tumors in mammography. 
Performance results were mixed. 
Additional data and future work could help clarify the potential utility of panoptic segmentation in this clinical task.

\clearpage
\printbibliography

\clearpage
\pagenumbering{gobble}

\section*{Supplementary Material}

\beginsupplement
\makeatletter
\renewcommand{\fnum@figure}{Fig. \thefigure}
\makeatother

\subsection*{S1. Ablation study on IoU thresholds}

\begin{figure}
\includegraphics[width=\columnwidth]{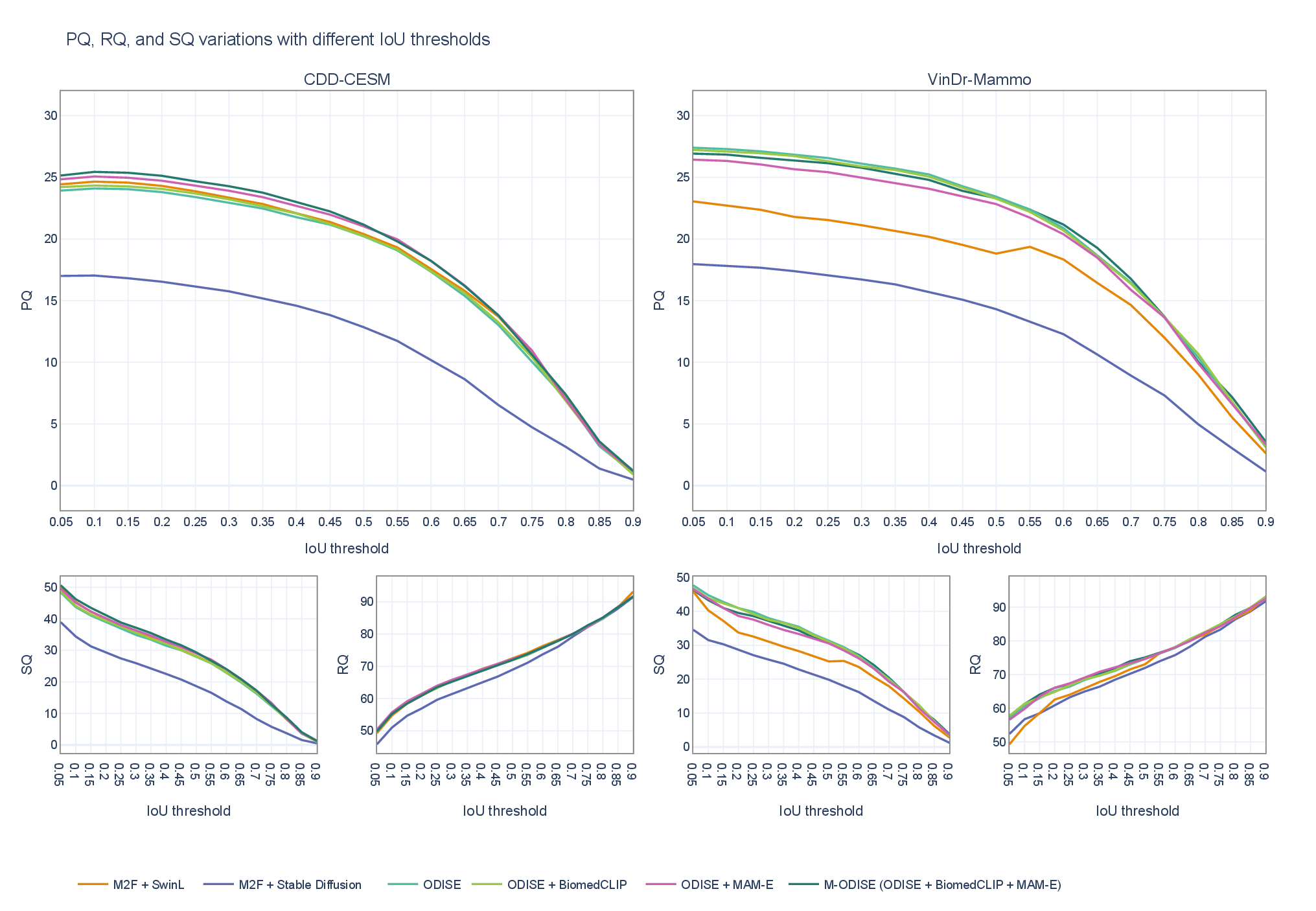}
\caption{Model performance across varying IoU thresholds. With a threshold value of 0.95, there were cases where no ground truth was matched; therefore, they are not included in the analysis.}
\label{fig: figS1}
\end{figure}

We analyzed the trade-off between RQ and SQ, which varies with the IoU threshold categorizing a prediction as positive. RQ measures the model's ability to detect and classify a lesion, while SQ defines how accurately the model can assign a segmentation mask to that finding. As the IoU threshold increases, RQ decreases because fewer predictions qualify as positive. Conversely, SQ increases with a higher threshold, as only more accurate predictions meet the criteria to be considered positive. The values of PQ, RQ, and SQ computed with different IoU thresholds are shown in Figure~\ref{fig: figS1}. For CDD-CESM, the optimal threshold is 0.1 for all models, while for VinDr-Mammo, this value is 0.05.

\subsection*{S2. Qualitative assessments}

Below, on Figure~\ref{fig: figS2}, we provide more qualitative visualizations of M-ODISE predictions.

\begin{figure}
\centering
\includegraphics[width=\columnwidth, keepaspectratio]{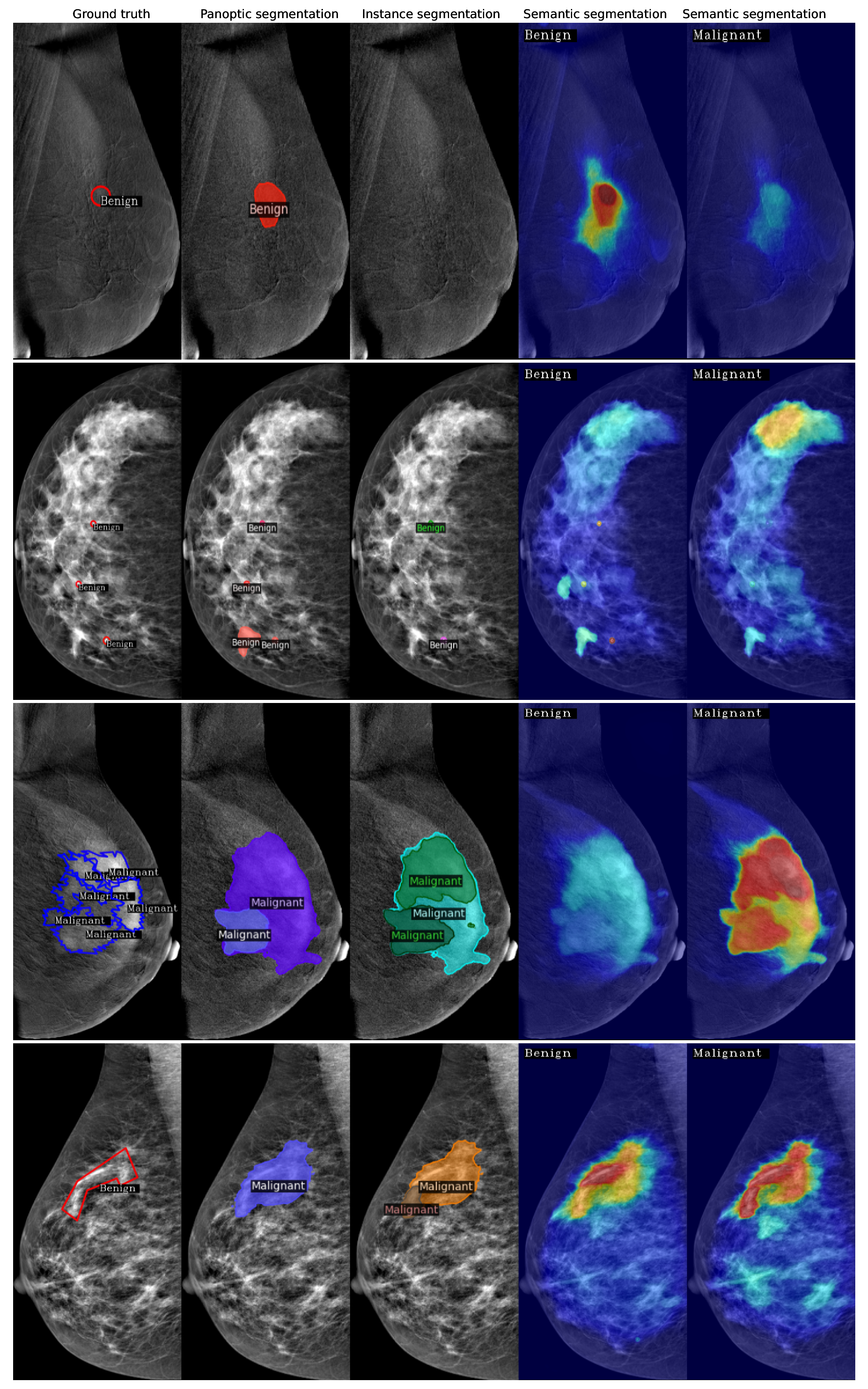}
\caption{
Qualitative visualization of M-ODISE predictions on the CDD-CESM dataset. The ground truth mask, panoptic segmentation mask, instance segmentation mask, and semantic heatmap are presented for four samples. 
}

\label{fig: figS2}
\end{figure}

\end{document}